\documentclass[letterpaper, 10pt,  journal, final, twocolumn]{IEEEtran}
\usepackage{graphicx}
\usepackage{amsmath,amsfonts}
\usepackage{color}
\usepackage{flushend}
\usepackage{balance}
\usepackage{amsthm}
\usepackage{cases}
\usepackage{multirow}
\usepackage{setspace}
\usepackage{empheq}
\usepackage{caption}
\usepackage{notoccite}
\usepackage{ulem}

\usepackage{booktabs}

 \newcommand{\blue}{\color{blue}}

\begin{document}

\title{{\small\textbf{\blue IEEE/MTS Global OCEANS 2021, San Diego - Porto}}\\
\Huge{A Non-uniform Sampling Approach for Fast and Efficient Path Planning}
\thanks{$^\dag$Dept. of Electrical and Computer Engineering, University of Connecticut, Storrs, CT 06269, USA.}
\thanks{$^{\star}$ Corresponding Author (email id: james.wilson@uconn.edu)}
\thanks{\copyright  2021 IEEE.  Personal use of this material is permitted.  Permission from IEEE must be obtained for all other uses, in any current or future media, including reprinting/republishing this material for advertising or promotional purposes, creating new collective works, for resale or redistribution to servers or lists, or reuse of any copyrighted component of this work in other works.}
}
\author{ \begin{tabular}{ccc}
{James P. Wilson$^\dag$$^\star$} & {Zongyuan Shen$^\dag$} & {Shalabh Gupta$^\dag$}\\
\end{tabular}
\vspace{-20pt}
}

\maketitle

\begin{abstract}
    In this paper, we develop a non-uniform sampling approach for fast and efficient path planning of autonomous vehicles. The approach uses a novel non-uniform partitioning scheme that divides the area into obstacle-free convex cells. The partitioning results in large cells in obstacle-free areas and small cells in obstacle-dense areas. Subsequently, the boundaries of these cells are used for sampling; thus significantly reducing the burden of uniform sampling. When compared with a standard uniform sampler, this smart sampler significantly 1)  reduces the size of the sampling space while providing completeness and optimality guarantee, 2) provides sparse sampling in obstacle-free regions and dense sampling in obstacle-rich regions to facilitate faster exploration, and 3) eliminates the need for expensive collision-checking with obstacles due to the convexity of the cells. This sampling framework is incorporated into the RRT* path planner. The results show that RRT* with the non-uniform sampler gives a significantly better convergence rate and smaller memory footprint as compared to RRT* with a uniform sampler.
    
\end{abstract}

\begin{IEEEkeywords}
non-uniform sampling; autonomous vehicles; sampling-based algorithms; near-optimal path planning
\end{IEEEkeywords}

\thispagestyle{empty}

\section{Introduction}
\label{sec:intro}

Autonomous vehicles are becoming increasingly useful and cost-effective for a variety of tasks in many scientific expeditions. Specifically, autonomous underwater vehicles (AUVs) have been extensively used for exploration~\cite{Song2018_eps*,SG19}, data collection (e.g., ocean salinity and temperature), ship hull cleaning \cite{Nasssiraei2012_HullCleaning}, underwater pipeline and cable monitoring \cite{Bagnitsky2011_pipelines}, marine wildlife monitoring \cite{Robbins2006_wildlife}, seabed mapping~\cite{palomeras2018,Shen2016_3DRecon,Shen2017_3DCTCPP,shen2020ct,Shen2019_DubinsCPP,Shen2020_EnergyCPP}, oil spill cleaning~\cite{Song2013_Oil},
mine hunting~\cite{Sariel2008_Mines,Mukherjee2011_SonarDet}, and other marine research. Despite recent advances, their autonomy is limited. AUV missions might require on-demand path synthesis in unknown or dynamic environments \cite{Garau2009_AUVA*,Mittal2020_DubinsDrift}. The autonomy of AUVs is determined by the proficiency in which they can replan paths as new information becomes available \cite{Zeng2015_AUVSurvey}. It is thus of practical importance for path planners to be computationally efficient and robust when constructing the cost-minimizing paths.

A review of path planning methods for AUVs is presented in \cite{Zeng2015_AUVSurvey}. In general, path planning can be divided into two categories: grid-based and sample-based. Grid-based methods, such as A* \cite{lavalle2006}, discretize the configuration space and search for the optimal solution; however, the solution quality depends on the grid resolution and suffers from the curse of dimensionality. On the other hand, sample-based methods, such as RRT* \cite{Karaman2011_RRT*,Wilson2020_TStarLite}, are becoming increasingly popular for on-demand motion planning since they can 1) find feasible solutions quickly in high-dimensional spaces and 2) approach the optimal solution as the number of samples increases. However, while these methods perform well in relatively open spaces, the uniform samplers often struggle with narrow passages since the chance of placing a sample within the passage with a collision-free connection is very small \cite{Sun2005_PRMNarrow}. Although a feasible path is eventually found, it requires a large number of samples and computationally expensive collision checks, thus providing a slow convergence rate.

To overcome the limitations of sample-based approaches, researchers have recently focused on sampling only the \textit{critical regions} \cite{Ichter2020_CriticalRegion,Molina2020_CriticalRegion}. These regions are often the entrances of a passage connecting two larger areas (e.g., doorways). The idea is that by biasing a large number of samples to these regions, sample-based planners can quickly find a feasible path. Furthermore, it encourages sparse sampling in obstacle-free regions, thus facilitating faster convergence. The identification of these critical regions has mostly been achieved using deep learning models trained on supplied examples of the shortest path \cite{Wang2020_NeuralRRT*,Qureshi2021_MPNet,Kumar2019_LEGO}. While the results are promising, these models require large and diverse datasets for training and are not guaranteed to find these regions. Furthermore, these approaches might not scale or be robust in new environments.

In this paper, we present a new path planning framework that combines the best features of grid-based and sample-based path planners for intuitively identifying, sampling and planning along the critical regions. These regions are identified using a new method inspired by renormalization group theory \cite{goldenfeld2018,wilson1971,Mittal2019_RGT}. This method partitions the region which creates non-overlapping large convex cells in obstacle-free spaces, and small convex cells in obstacle-dense regions. The critical regions are the obstacle-free boundaries of these cells. Then, a feasible path is found quickly by sampling only in these regions using a sample-based path planner. In particular, the convexity of the cells joining critical regions eliminates the need for collision checking, which significantly reduces the computation time. Finally, once a feasible path is found, a local search is performed to find the (near)optimal solution.

The rest of this paper is organized as follows. Section~\ref{sec:prob} formulates the path planning problem for AUVs. Section~\ref{sec:method} presents the smart sampling procedure and the path planning algorithm. Section~\ref{sec:results} shows the results on a simulated scenario, and Section~\ref{sec:conclusion} concludes this paper with recommendations for future work.

\begin{figure*}[t!]
    \centering
        \centering
        \includegraphics[width=1\textwidth]{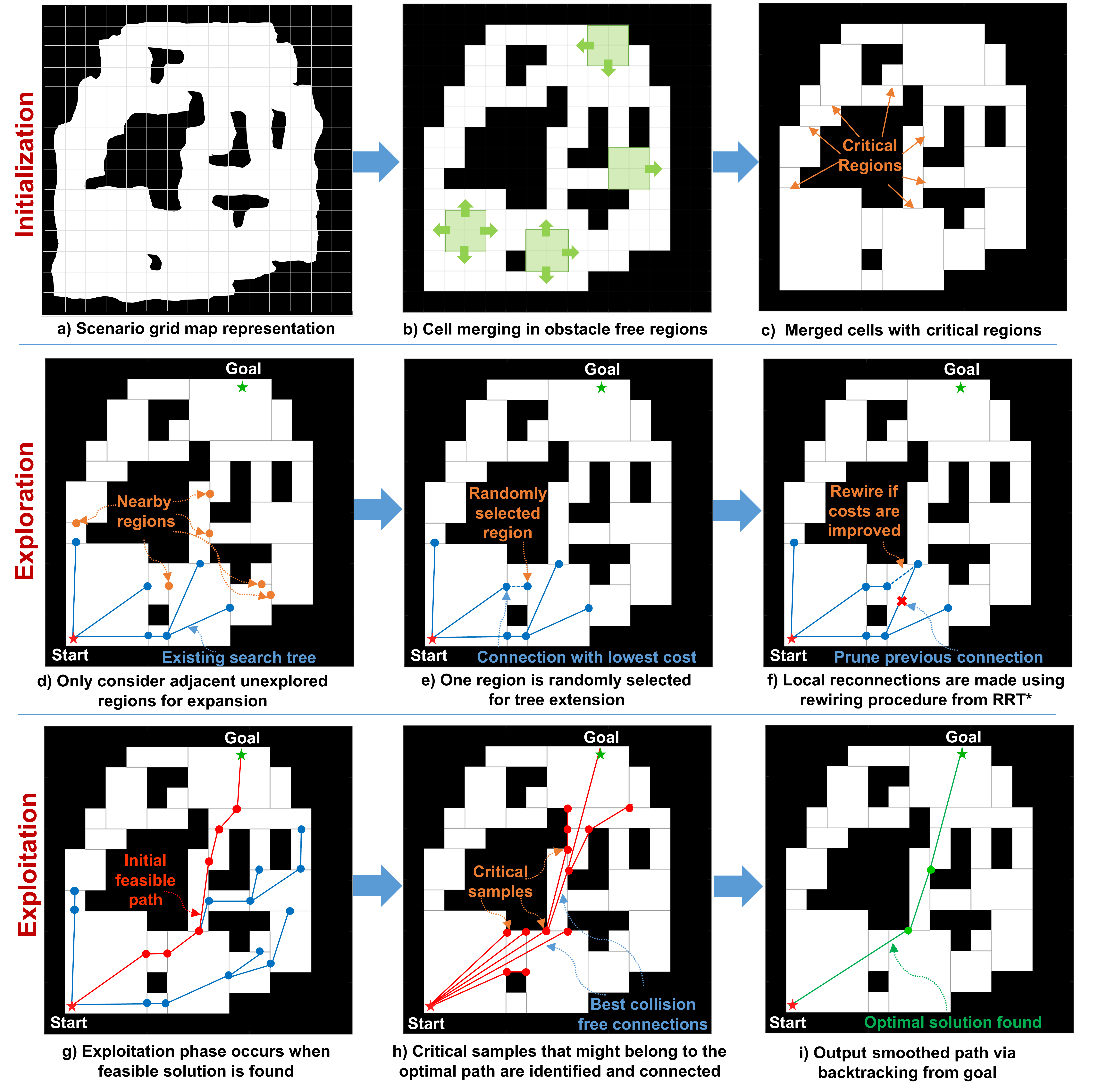}
         \caption{
         Non-uniform sampling for fast and efficient {RRT$^*$-based} path planning. 
         }
         \label{fig:main}
         \vspace{-10pt}
\end{figure*}

\section{Problem Formulation}
\label{sec:prob}

Let $\mathcal{A}\subset \mathbb{R}^2$ be the 2D search area composed of free space $\mathcal{A}_{f}$ and obstacle space $\mathcal{A}_{o}$, where $\mathcal{A}=\mathcal{A}_f\cup\mathcal{A}_o$, $\mathcal{A}_f\cap\mathcal{A}_o=\emptyset$, and $\mathcal{A}_f$ is a connected space. The position of the AUV is denoted as $\mathbf{p}=(x,y)\in\mathcal{A}_f$. Define $\Gamma$ as the set of all collision-free paths from the start position $\mathbf{p}_{start}$ to the goal position $\mathbf{p}_{goal}$. For every feasible path $\gamma\in\Gamma$, 
its length is:


\begin{equation}
    J(\gamma)=\int_\gamma ds
\end{equation}
The objective is to find the path $\gamma^*$ s.t. $J(\gamma^*)\leq J(\gamma), \forall \gamma\in\Gamma$.

\section{Smart Sampling Approach}
\label{sec:method}

The basic idea is outlined in Figure~\ref{fig:main} and is described in three phases: initialization, exploration, and exploitation. The critical regions for non-uniform sampling are found during the initialization phase. First, the known environment map is represented as a high-resolution uniform grid. Then, this map is converted into a coarse non-uniform grid by merging large groups of obstacle-free cells together. The boundaries of these merged cells form the \textit{critical regions} for sampling. Next, in the exploration phase, a modified RRT* algorithm efficiently explores the space by only sampling these critical regions. To facilitate faster exploration, a region is considered \textit{explored} when it is first sampled, and only unexplored adjacent regions are considered in the next iteration. Connections between adjacent regions are guaranteed to be collision-free since the adjoining grid cell is a convex obstacle-free space. Finally, once a feasible solution is found, the (locally) optimal path is quickly obtained during the exploitation phase. From visibility graph theory, the corner points around obstacles are known to be a part of the optimal solution \cite{Zafar2021_LTA}. As such, the end points of the critical regions along the feasible path are identified. Then, a graph with the shortest paths to each of these points is created, and the (locally) optimal solution is obtained.

\subsection{Initialization}
The critical regions for non-uniform sampling are identified in the initialization phase. In order to find these regions, the search space $\mathcal{A}$ is partitioned into $N\times N$ uniform grid $\mathcal{C}$ with pairwise disjoint cell interiors, i.e., $\mathcal{C}=\{c(i,j)\subset\mathbb{R}^2:i,j=1,\dots,N\}$, such that $c^o(i,j)\cap c^o(k,l)=\emptyset, \forall i\neq k$ and $j\neq l$, and $\bigcup_{i,j}c(i,j)= \mathcal{A}$, where $c(i,j)$ is a cell located at position $(i,j)$ of the grid and $^o$ denotes the interior of a cell. A cell $c(i,j)$ is denoted as an obstacle cell $c_o(i,j)$ if it is partially or fully occupied by any obstacle; otherwise, $c(i,j)$ is denoted as a free cell $c_f(i,j)$. Figure~\ref{fig:main}-a shows an example of partitioning an obstacle-rich scenario. 

Once the obstacle and free cells are obtained from the grid partitioning, the free cells are expanded and merged together if they belong to the same convex obstacle-free group, as shown in Figure~\ref{fig:main}-b. The merging procedure is as follows. To create merged cell group $C_m$, where $m=1,2,\dots,M$ and $M$ is the total number of groups (determined once all free cells are merged), select any free cell $c_f(i,j)$, that has not been merged. Check all of its immediate eight neighboring cells $c(i-1,j-1),c(i-1,j),c(i-1,j+1),\dots,c(i+1,j+1)$, and merge the cells into $C_m$ that create the largest obstacle-free rectangular group of unmerged cells including $c_f(i,j)$. Keep expanding along the boundaries of cell group $C_m$ in this manner until no more cells can be merged in any direction. Then, pick any unmerged free cell $c_f(i,j)$ and create the next cell group $C_{m+1}$ in the same manner. Repeat until all free cells belong to a cell group.

Once the cell groups $C_1,\dots,C_M$ are created, the critical regions for sampling are identified. Note that the cell groups: 1) have disjoint interiors, i.e., $C^o_m\cap C^o_n = \emptyset$, 2) include all free cells, i.e., $\bigcup_m C_m=\bigcup_{i,j}c_f(i,j)$, and 3) are convex since they are rectangular. The sampling regions are defined as the (obstacle-free) boundaries (i.e., lines) between all pairs of neighboring cell groups. We denote any such boundary line between neighboring cells $C_m$ and $C_n$ as $r_{mn}$. Note that $r_{mn}$ and $r_{nm}$ are identical. Figure~\ref{fig:main}-c shows the critical regions.

\subsection{Exploration}

During the exploration phase, the search tree is grown using a modified RRT* path planner until a feasible solution to the goal is found. The RRT* algorithm consists of six main functions: \textit{sampling}, \textit{distance}, \textit{nearest neighbor}, \textit{nearby vertices}, \textit{collision check}, and \textit{local steering}. In order to utilize the critical regions for smart sampling, only the \textit{sampling} and \textit{nearby vertices} are updated. Additionally, we create one new function called \textit{nearby regions}, which identifies all regions close to the search tree. All other functions are the same. For brevity's sake, we refer the reader to \cite{Karaman2011_RRT*} for more details on RRT*. It should noted that the \textit{distance} function used in this paper is the Euclidean distance since the objective is to find the shortest path. 

\vspace{6pt}
\subsubsection{Nearby Regions} Given an existing search tree $T=(V,E)$, where $V$ is the set of nodes denoting the samples and $E$ is the set of connections between the samples, the set of nearby regions is:
\small
\begin{equation}
    \begin{split}
          \text{NearbyRegions}(T) & =\{r_{mn} : \\
         & v \in r_{mj} \lor v \in r_{nj} \lor 
         v\in C^o_m \lor v \in C^o_n, \\
       & \forall v\in V, v\notin r_{mn}, j\neq m,n  \}
    \end{split}
\end{equation}
Note that  $v\in C^o_m$  or $C^o_n$ if $C^o_m$ or $C^o_n$ contain the start or the goal. \normalsize In other words, the nearby regions to search tree $T$ are the regions that have not yet been sampled but share a cell group with an already sampled region. An example of the nearby regions for a given search tree are shown in Figure~\ref{fig:main}-d.

\vspace{6pt}
\subsubsection{Sampling} Given the set NearbyRegions($T$), a nearby region is randomly selected for tree expansion according to a discrete uniform distribution, i.e., if there are $N_{nr}$ nearby regions, the probability of selecting any region is $1/N_{nr}$. Then, a new node $v_{new}$ is sampled in the center of this region. It is then connected to its nearest neighbor in search tree $T$ in the same manner as RRT*. An example is shown in Figure~\ref{fig:main}-e. Note that this sampling paradigm limits the number of samples per region to one in order to facilitate faster exploration.

\vspace{6pt}
\subsubsection{Nearby Vertices} Given a search tree $T=(V,E)$ and sample $\mathbf{p}=(x,y)\in\mathbb{R}^2$ lying on some region $r_{mn}$, the set of nearby vertices $\{v_{near}\}$ consists of vertices that either lie 1) on regions that adjoin either cell group $C_m$ or $C_n$ with some other cell group $C_i$, or 2) within the interiors of $C^o_m$ or $C^o_n$:
\small
\begin{equation}
\begin{split}
    \text{NearbyVertices}(T,\mathbf{p}) & =\{ v_{near}\in V : \\  
    &  v_{near}\in r_{mi} \lor v_{near}\in r_{ni}  \lor\\
    & v_{near}\in C^o_m \lor v_{near}\in C^o_n , \\
    & \; \forall i = 1,\dots,M,\; i\neq n,m,  \; \mathbf{p}\in r_{mn}\}
\end{split}
\end{equation}
\normalsize If instead point $\mathbf{p}$ lies inside the interior of $C^o_m$ and not on any region $r_{mj}$ , the nearby vertices are:
\small
\begin{equation}
\begin{split}
    \text{NearbyVertices}(T,\mathbf{p}) & = \{ v_{near}\in V: \\
    & v_{near}\in C^o_m \lor v_{near}\in r_{mj}, \\
      & \forall j=1,\dots,M\}
\end{split}
\end{equation}
\normalsize Note that the only points $\mathbf{p}$ that lie in the interior of the cell groups are $\mathbf{p}_{start}$ and $\mathbf{p}_{goal}$. The function NearbyVertices is used to find the the set of nearest neighbors for connecting a newly sampled node to the tree and performing a local rewiring of the tree as needed in the same manner as RRT*. An example is shown in Figure~\ref{fig:main}-f.

\begin{figure*}[t!]
    \centering
        \centering
        \includegraphics[width=.85\textwidth]{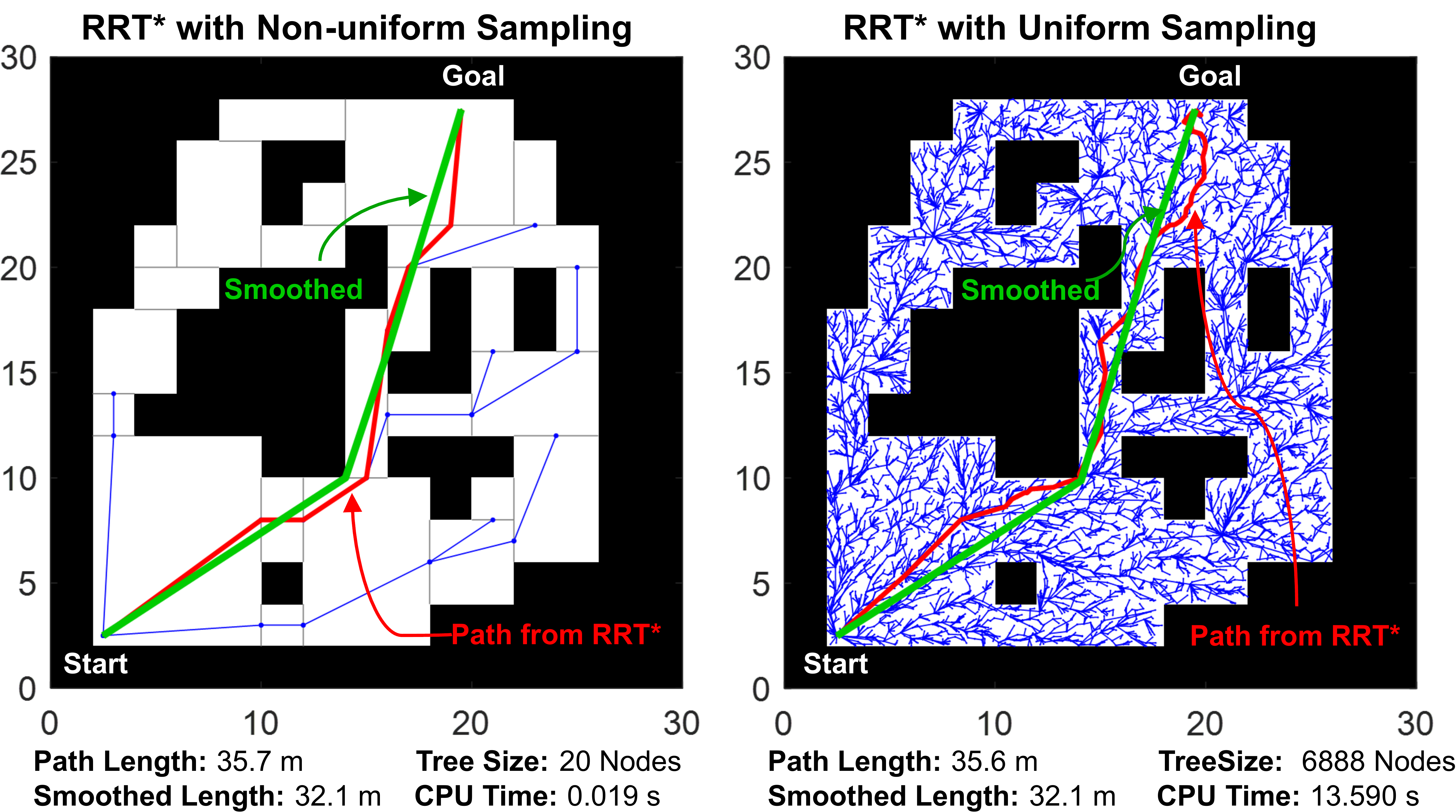}
         \caption{
         Performance comparison of RRT* with the non-uniform sampler (left) and RRT* with the uniform sampler (right).
         }
         \label{fig:results}
         \vspace{-10pt}
\end{figure*}

\subsection{Exploitation}
Once a feasible path is found, as shown in Figure~\ref{fig:main}-g, a search begins for the (locally) optimal solution within the homotopy class of this path. From visibility graph theory \cite{Zafar2021_LTA}, the corner points of obstacles create the sufficient set of points that could belong to the shortest path. As such, new samples are made at the end points of each region $r_{mn}$ that the feasible solution passes through. This set of samples contains at least all of the corner points along the feasible solution. Then, a new search tree is created using only these samples with the shortest collision-free connections made to the start. An example is shown in Figure~\ref{fig:main}-h. Note that in this step, collision-checking is necessary since connections between two samples can traverse several cell groups. Finally, the (locally) optimal path is obtained from this search tree by backtracking from the goal to the start, as shown in Figure~\ref{fig:main}-i.

\section{Results and Discussion}
\label{sec:results}
In this section, the performance of RRT* with the non-uniform sampler is compared against that of the standard RRT* with a uniform sampler. The scenario considered is a $30m\times30m$ map populated with several obstacles, as shown in Figure~\ref{fig:results}. For  non-uniform sampling, the scenario is uniformly partitioned into cells of size $2m\times2m$, resulting in a grid map of $15\times15$ cells. 
For the RRT* planner with the uniform sampler, the maximum connection distance between samples is set to be $5m$. The simulation was carried out in MATLAB using the RRT* path planner in the Navigation Toolbox on a Windows 10 machine with an Intel Core-i7 7700 CPU and 32GB of RAM.

Figure~\ref{fig:results} shows the results generated by RRT* using both the non-uniform sampler (left) and the uniform sampler (right). Each plot shows: 1) the search tree in blue, 2) the best feasible path found from the RRT* path planner in red, and 3) the smoothed path in green. The benefits of the non-uniform sampler are evident since the identification of the critical regions has enabled the RRT* planner to quickly and efficiently cover the search space. The feasible path found goes directly to the goal through the narrow corridors between the obstacles in the center of the map, providing a smoothed path length of $32.1m$. Moreover, this path is found with only $20$ samples in the search tree and a total computation time of $0.019$ seconds. This low computation time is achieved since 1) the  non-uniform sampler greatly reduces the size of the sampling space, and 2) no collision-checking is required during the exploration phase. 

On the other hand, RRT* with the uniform sampler struggles in the obstacle-rich center of the map and thus needs a significantly longer computation time of $13.59$ seconds (about 16,000 iterations) and a much larger tree size of $6888$ nodes to find a path of the same solution quality of  path length of $32.1m$ as compared to RRT* with the non-uniform sampler. Overall, the proposed non-uniform sampling approach shows significant promise over the uniform sampling, providing an intuitive way to identify these critical regions and efficiently find the shortest path.

\section{Conclusions and Future Work}
\label{sec:conclusion}
In this paper, we developed a novel non-uniform sampler that intuitively identifies the critical regions for fast and efficient path planning. This is achieved by developing a new partitioning method that creates large convex cells in obstacle-free spaces and smaller convex cells in obstacle-dense regions. Then, we illustrate how to sample and search for a path along these regions in the RRT* framework. Specifically, non-uniform sampling speeds up the search significantly by dramatically reducing the size of the sampling space and systematically connecting samples in a way that does not require expensive collision checking during exploration. Compared to RRT* with a uniform sampler, this approach provides the same solution quality in  significantly less computation time and memory footprint. Compared to other methods that identify critical regions using deep learning, this approach guarantees complete coverage of the configuration space without the necessity of uniform sampling.

Future work will include in-depth theoretical analysis of the non-uniform sampler with proofs for completeness and optimality. The partitioning scheme will be updated to search for the critical regions incrementally instead of requiring the entire map be partitioned a-priori. Informed subsets \cite{Gammell2018_informed} will also be incorporated into this non-uniform sampling framework to provide an even faster convergence rate. Finally, this work will be extended to provide these nonuniform sampling benefits to multi-speed non-holonomic vehicles where both travel time and collision risk \cite{song2018} are considered in the cost \cite{Wilson2019_GMDM,Wilson2020_TStarLite}.
\balance
\bibliographystyle{ieeetr}
\bibliography{References}

\end{document}